\documentclass[conference]{IEEEtran}
\usepackage{cite}
\usepackage{url}
\usepackage{hyperref}
\usepackage{authblk}
\usepackage{enumitem}
\setlist{nosep}
\usepackage{amsmath}
\usepackage{amssymb}
\usepackage[noend]{algpseudocode}
\usepackage{booktabs}
\usepackage{multirow}
\usepackage{tabularx}
\usepackage{xcolor}
\usepackage{color, colortbl}
\usepackage{epsfig}
\usepackage{graphicx}
\usepackage{subcaption}
\DeclareMathOperator*{\argmax}{argmax}

\definecolor{lightgray}{gray}{0.9}

\def\onedot{.}
\def\eg{\emph{e.g}\onedot} 
\def\ie{\emph{i.e}\onedot} 
\def\cf{\emph{cf}\onedot}

\def\etal{\emph{et al}\onedot}

\begin{document}

\title{Rethinking Adversarial Examples for \\ Location Privacy Protection}

\author[*, 1]{Trung-Nghia Le
% \thanks{These authors have equal contributions}
}
\author[*, 2]{Ta Gu} 
\author[*, 1]{Huy H. Nguyen}
\author[1, 3]{Isao Echizen}

\affil[1]{National Institute of Informatics, Japan}
\affil[2]{University of Electronic Science and Technology of China, China}
\affil[3]{University of Tokyo, Japan}
\affil[*]{These authors have equal contributions}

% \author{\IEEEauthorblockN{Michael Shell}
% \IEEEauthorblockA{School of Electrical and\\Computer Engineering\\
% Georgia Institute of Technology\\
% Atlanta, Georgia 30332--0250\\
% Email: http://www.michaelshell.org/contact.html}
% \and
% \IEEEauthorblockN{Homer Simpson}
% \IEEEauthorblockA{Twentieth Century Fox\\
% Springfield, USA\\
% Email: homer@thesimpsons.com}
% \and
% \IEEEauthorblockN{James Kirk\\ and Montgomery Scott}
% \IEEEauthorblockA{Starfleet Academy\\
% San Francisco, California 96678--2391\\
% Telephone: (800) 555--1212\\
% Fax: (888) 555--1212}}

% \author{
% \IEEEauthorblockN{
% Trung-Nghia Le\IEEEauthorrefmark{1},
% Ta Gu\IEEEauthorrefmark{2},
% Huy H. Nguyen\IEEEauthorrefmark{1},
% Isao Echizen\IEEEauthorrefmark{1, 3}
% }
% \IEEEauthorblockA{\IEEEauthorrefmark{1}National Institute of Informatics, Japan\\ 
% Email: \{ltnghia,nhhuy,iechizen\}@nii.ac.jp}
% \IEEEauthorblockA{\IEEEauthorrefmark{2}University of Electronic Science and Technology of China, China\\
% Email: 2018190607036@std.uestc.edu.cn}
% \IEEEauthorblockA{\IEEEauthorrefmark{3}University of Tokyo, Japan}
% }

% conference papers do not typically use \thanks, and this command is locked out in conference mode. If needed, such as for the acknowledgment of grants, issue a \IEEEoverridecommandlockouts after \documentclass

% used for special paper notices
%\IEEEspecialpapernotice{(Invited Paper)}

% make the title area
\maketitle

% As a general rule, do not put math, special symbols, or citations in the abstract

\begin{abstract}
We have investigated a new application of adversarial examples, namely location privacy protection against landmark recognition systems. We introduce mask-guided multimodal projected gradient descent (MM-PGD), in which adversarial examples are trained on different deep models. Image contents are protected by analyzing the properties of regions to identify the ones most suitable for blending in adversarial examples. We investigated two region identification strategies: class activation map-based MM-PGD, in which the internal behaviors of trained deep models are targeted; and human-vision-based MM-PGD, in which regions that attract less human attention are targeted. Experiments on the Places365 dataset demonstrated that these strategies are potentially effective in defending against black-box landmark recognition systems without the need for much image manipulation.
\end{abstract}

% no keywords

% For peer review papers, you can put extra information on the cover page as needed:
% \ifCLASSOPTIONpeerreview
% \begin{center} \bfseries EDICS Category: 3-BBND \end{center}
% \fi
%
% For peer-review papers, this IEEEtran command inserts a page break and creates the second title. It will be ignored for other modes.
\IEEEpeerreviewmaketitle

%%%%%%%%% BODY TEXT
\section{Introduction}
\label{sec:intro}

The massive application of powerful computer vision systems has enabled machines to perceive the world. Incremental learning using a tremendous amount of data obtained from the Internet and surveillance cameras has enabled intelligent systems to track and learn the behaviors of a large population~\cite{Chris-Sensetime_surveillance, Weyand-ECCV2016}. Person recognition~\cite{Chris-Sensetime_surveillance, ClearviewAI-2022} and landmark detection~\cite{Weyand-ECCV2016} systems are usually deployed as cloud-based services for use in identifying people and tracking their locations and movements. For example, SenseTime, a Chinese AI software company, deploys a massive video surveillance network with biometric facial recognition across 1.9 million sites~\cite{Chris-Sensetime_surveillance}. Likewise, Clearview AI Inc., a U.S.-based company, obtained more than 20 billion facial images worldwide from social media platforms and other sites on the Internet and used them to create a database for facial recognition~\cite{ClearviewAI-2022}. Google as well developed a system trained on millions of images; their PlaNet system~\cite{Weyand-ECCV2016} can determine the location where a photo was taken by using only its pixels with more precision than even well-traveled humans.

Such systems can be trained using images and videos collected from social networks without the explicit consent of the people in the images and videos, leading to a serious privacy issue. In addition, social networks can utilize tools that recognize locations in shared images and use them to track users’ locations without user authorization. Several companies have violated the EU's privacy rules by acquiring personal information and using it to build highly detailed online profiles. For example, Google has violated children’s privacy by illegally extracting children’s data from YouTube videos without their parents’ consent and using it to profit by targeting them with advertisements~\cite{Natasha-Google}. Biometric images and videos of Instagram users are illegally collected by automatically scanning the faces of people pictured in other users' posts even if they do not use Instagram and do not agree to the terms of service and then using the images for targeted advertising~\cite{Aaron-Instagram}. The personal data of millions of Facebook users were collected without consent by Cambridge Analytica, a British consulting firm, mainly for use in political advertising~\cite{Nicholas-Analytica}. 

\begin{figure}[t!]
    \centering
    \includegraphics[width=\linewidth]{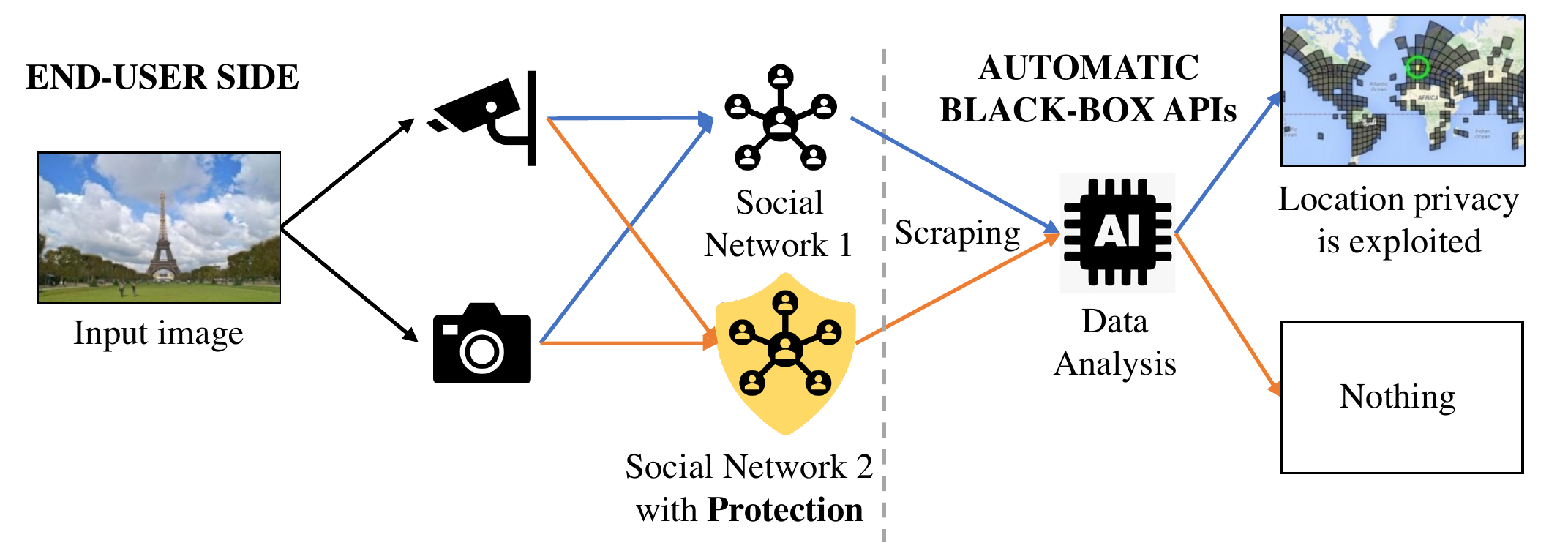}
    \caption{Location privacy protection scenarios in the cyber world. We investigated the ability of our MM-PGD method to prevent the location of users in images from being recognized and tracked by online landmark recognition systems.}
    \label{fig:privacy_protection}
\end{figure}

Privacy violation issues require governments and companies to establish policies to prevent unauthorized surveillance and tracking of user data~\cite{Sobel-Hiq2020}. The European Union (EU) passed a law on data protection and privacy in 2018, the General Data Protection Regulation (GDPR)\footnote{\url{https://gdpr-info.eu}}, that requires companies to implement safeguards to protect personal data. Google subsequently made changes to protect children’s privacy on YouTube~\cite{Natasha-Google}, and Facebook banned automated data collection on their websites without Facebook's express written permission\footnote{\url{https://www.facebook.com/apps/site\_scraping\_tos\_terms.php}}. In addition, Clearview AI Inc.~\cite{ClearviewAI-2022} was recently fined more than $\pounds7.5$ million and ordered to delete all UK data, including images and personal information.

Users of social networks (\ie, end-users) also need tools to protect their privacy by preventing the contents of their shared images and videos from being automatically scraped (\cf~Fig.~\ref{fig:privacy_protection}). End-users mainly need a tool that protects images uploaded to a social network so that image geolocation systems (\eg, PlaNet~\cite{Weyand-ECCV2016}) cannot analyze them to determine where they were taken. The protection tools should be able against unknown systems in order to be widely used for arbitrary social media platforms. In addition, the protected images should appear natural to human vision. 

Adversarial examples have been developed to hinder convolutional neural networks (CNNs). These methods aim to spoof object classification networks (\eg, deep models) in the digital world by applying adversarial examples directly to the digital images or videos~\cite{Goodfellow-ICLR2014, Shamsabadi-CVPR2020, Bhattad-ICLR2020}. These adversarial examples can be applied to uploaded images as shown in Fig.~\ref{fig:privacy_protection} to prevent cloud-based recognition APIs from exploiting information from the images when they are shared on social networks.

We investigated adversarial example-based solutions that prevent end-user content from being automatically scraped and analyzed when it is shared on social networks. Although adversarial examples were originally developed for attack purposes, we devised a new application for them: \textit{location privacy protection}. To this end, we extended the projected gradient descent (PGD) method~\cite{Bubeck-arxiv2014} to \textbf{M}ask-Guided \textbf{M}ultimodal \textbf{P}rojected \textbf{G}radient \textbf{D}escent (MM-PGD) in which adversarial examples are trained on different deep models to enhance their transferability. We also devised a means to identify the important regions (\ie, landmarks or other features) in images to enable them to be protected by embedding adversarial examples. This problem is more challenging than attacking object classification systems because we need to analyze the properties of the regions in each image in order to exploit the regions containing the most features of the scene. We investigated two strategies for identifying the regions that need to be protected: class activation map (CAM)-based MM-PGD, which targets the internal behavior of trained deep models, and human vision (HV)-based MM-PGD, which targets regions that attract less human attention. Experiments on the Places365 dataset~\cite{Zhou-TPAMI2017} demonstrated the potential of our MM-PGD methods. Our MM-PGD is robust against black-box landmark recognition systems. Furthermore, created adversarial images have high quality and naturalness in human vision. We expect that our results will help in the development of solutions to defend against landmark recognition with minimal image degradation. 

Our contributions are as follows:
\begin{itemize}
\item We present a new application of adversarial examples, namely location privacy protection against landmark recognition systems.
\item We introduce mask-guided multimodal projected gradient descent (MM-PGD). Adversarial examples are trained on different deep models to improve their transferability.
\item We present two strategies for region identification, CAM-based MM-PGD and HV-based MM-PGD, that defend well against landmark recognizers and do not need much manipulation of the original images.
\item We define the top-$k$ protection rate (PR) and use it as a metric to measure the success of using adversarial examples against classifiers.
\end{itemize}

The remainder of this paper is organized as follows. Section~\ref{sec:related_work} summarizes related work on visual privacy protection. Next, Section~\ref{sec:proposed_method} presents our methods. Section~\ref{sec:experiments} reports the results of the evaluation and in-depth analysis of our methods. Finally, Section~\ref{sec:conclusion} summarizes the key points and mentions future work.

% Although conventional adversarial example methods can fool DNN-based classifiers, adversarial images generated from these methods are fragile and easily disabled by transformations and compression algorithms. Moreover, they manipulate entire images, resulting in unnatural images to the human vision. 

% https://arxiv.org/pdf/2107.01396.pdf
% https://arxiv.org/pdf/1801.02610.pdf
% https://arxiv.org/pdf/2006.12655.pdf
% https://arxiv.org/pdf/2107.01396.pdf
% https://arxiv.org/pdf/2203.05151.pdf
% https://github.com/LinQinLiang/SSAH-adversarial-attack
% 
% 

% AnonymousNet: Natural Face De-Identification with Measurable Privacy

% https://arxiv.org/pdf/1904.12620.pdf

% \begin{figure}[t!]
%     \centering
%     \includegraphics[width=\linewidth]{images/AE.jpg}
%     % \caption{Left: original image. Middle: adversarial perturbation found by PGD against ResNet50~\cite{He-CVPR2016}, size of perturbation is magnified 100 times to be more visible. Right: adversarial example result. \highlight{this figure is totally incorrect; the middle is gradient, not perturbation.}}
%     % \caption{Example of misclassification due to adversarial examples.}\footnote{\url{https://medium.com/swlh/gradient-based-adversarial-attacks-an-introduction-526238660dc9}}
%     \caption[ae]{Example of misclassification due to adversarial examples.\protect\footnotemark}
%     \label{fig:pgd}
% \end{figure}
% \footnotetext{https://medium.com/swlh/gradient-based-adversarial-attacks-an-introduction-526238660dc9}

% \section{Related Work}
\section{Visual Privacy Protection}
\label{sec:related_work}

% \highlight{in working}

% \subsection{Visual Privacy Protection}

Visual privacy protection prevents the data in images or videos that an individual wants to keep private from becoming available in the public domain. Private data includes the person's identity and sensitive information.

% https://eprints.kingston.ac.uk/id/eprint/31628/1/Florez-Revuelta-F-31628.pdf

People can hide their identity behind a virtual identity or in blind-vision texture to avoid being identified. For example, Li \etal~\cite{Li-TIFS2012} combined two fingerprints captured from different fingers to create a virtual fingerprint. In the authentication process, the virtual fingerprint is matched with the two original ones using a minutiae-based fingerprint matching algorithm. To protect privacy against unmanned aerial vehicles, Lee \etal~\cite{Lee-IEEEAccess2021} proposed transforming a person's face into a different face by using a generative adversarial network. In this “face-anonymizing drone patrol system,” each modified face part looks like the face of a person who does not exist. Frome \etal~\cite{Frome-ICCV2009} combined a sliding-window detector tuned for a high-recall low-precision operating point with a fast post-processing stage that is able to remove additional false positives for blurring faces in Google Street View. Chatzikyriakidis \etal~\cite{Chatzikyriakidis-ICIP2019} blended adversarial examples into facial images for face de-identification in order to fool automatic face recognition systems.

Sensitive information in images and videos (\eg, clothes, properties, and locations) can reveal the private lives of people, such as their routines, habits, and wealth. Therefore, such information must be protected from being captured. Harvey~\cite{Harvey-camoflash2010} developed an anti-paparazzi device that uses an array of high-power LEDs to produce a stream of light of over 12K lumen that blinds an optical camera lens. Frome \etal~\cite{Frome-ICCV2009} presented a system that automatically detects and blurs license plates for privacy protection in Google Street View. Treu \etal~\cite{Marc-CVPRW2021} overlaid adversarial textures on clothing regions to make everyone in an image undetectable. Unlike previous work, we aimed at preventing landmarks and locations in photos scraped from social networks from being recognized and tracked by landmark recognition systems.

% \subsection{Gradient-Based Adversarial Attack}

\section{Methodology}
\label{sec:proposed_method}

% https://openaccess.thecvf.com/content/ICCV2021/papers/Wang_Admix_Enhancing_the_Transferability_of_Adversarial_Attacks_ICCV_2021_paper.pdf

This section provides details of the investigated problem, followed by several gradient-based adversarial examples to which our method is most related. Then we introduce our multimodal adversarial examples.

\subsection{Problem Formulation}

% https://adversarial-ml-tutorial.org/adversarial_examples/
% https://engineering.purdue.edu/ChanGroup/ECE595/files/chapter3.pdf
% https://arxiv.org/pdf/1807.00051.pdf
% https://arxiv.org/pdf/2101.05639.pdf
% https://adversarial-ml-tutorial.org/introduction/

Given an image $x$, we aim to protect the image against a target deep classifier $C_{\theta}$ with loss function %$\ell(C_{\theta}(x),y)$
$\mathcal{L} ( x, y ; \theta )$. In particular, we aim to maximize the loss by solving an optimization problem:
\begin{equation}
\label{eq:ae1}
    % \widehat{x} = \argmax_{x}\ell(C_{\theta}(x),y),
    \widehat{x} = \argmax_{x}\mathcal{L} ( x, y ; \theta ),
\end{equation}
where $\widehat{x}$ denotes the adversarial example that is attempting to maximize the loss corresponding to the true predicted label $y$. We cannot optimize arbitrarily over $x$ but make small modifications to ensure that $\widehat{x}$ is close to the original input $x$. We do this by optimizing over the adversarial perturbation $\delta$:
\begin{equation}
\label{eq:ae2}
    % \widehat{x} = x + \argmax_{\left \| \delta \right \| \leq \epsilon}\ell(C_{\theta}(x + \delta),y),
    \widehat{x} = x + \argmax_{\left \| \delta \right \| \leq \epsilon}\mathcal{L} ( x + \delta, y ; \theta ),
\end{equation}
where $\epsilon$ is the perturbation magnitude boundary. In this paper, we consider the common perturbation in the $\ell_\infty$-ball, in which the norm $\left \| \delta \right \|_\infty=\max_i |\delta_i|$ is used.

% In following sections, we simplify the loss function $\ell(C_{\theta}(x,y)$ by $\mathcal{L} ( x, y ; \theta )$ and use it in all equations. Therefore, Eq.~\ref{eq:ae2} is formulated as:
% \begin{equation}
%     \widehat{x} = \argmax_{x} \mathcal{L} ( x, y ; \theta ).
% \end{equation}
% \begin{equation}
%     \widehat{x} = x + \argmax_{\left \| \delta \right \| \leq \epsilon}\mathcal{L} ( x + \delta, y ; \theta ).
% \end{equation}

\begin{figure}[t!]
    \centering
    \includegraphics[width=\linewidth]{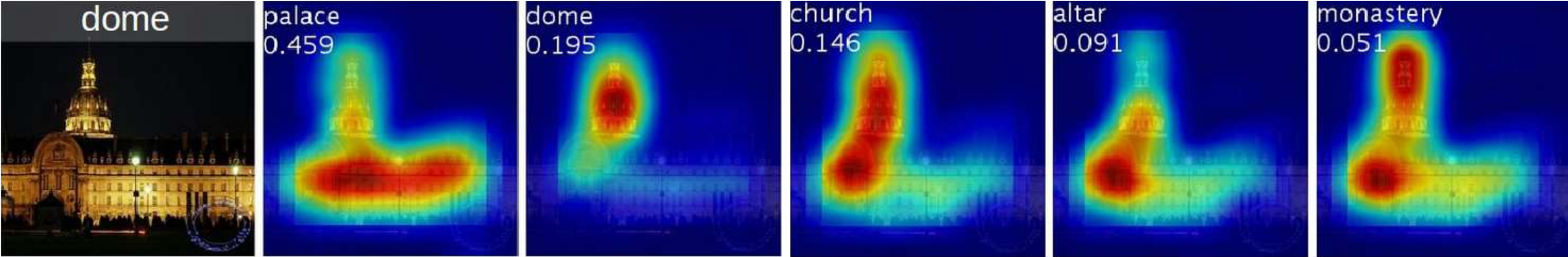}
    \caption{Class activation map highlights regions of the top-5 predicted classes.}
    \label{fig:cam}
\end{figure}

\begin{figure}[t!]
    \centering
    \includegraphics[width=\linewidth]{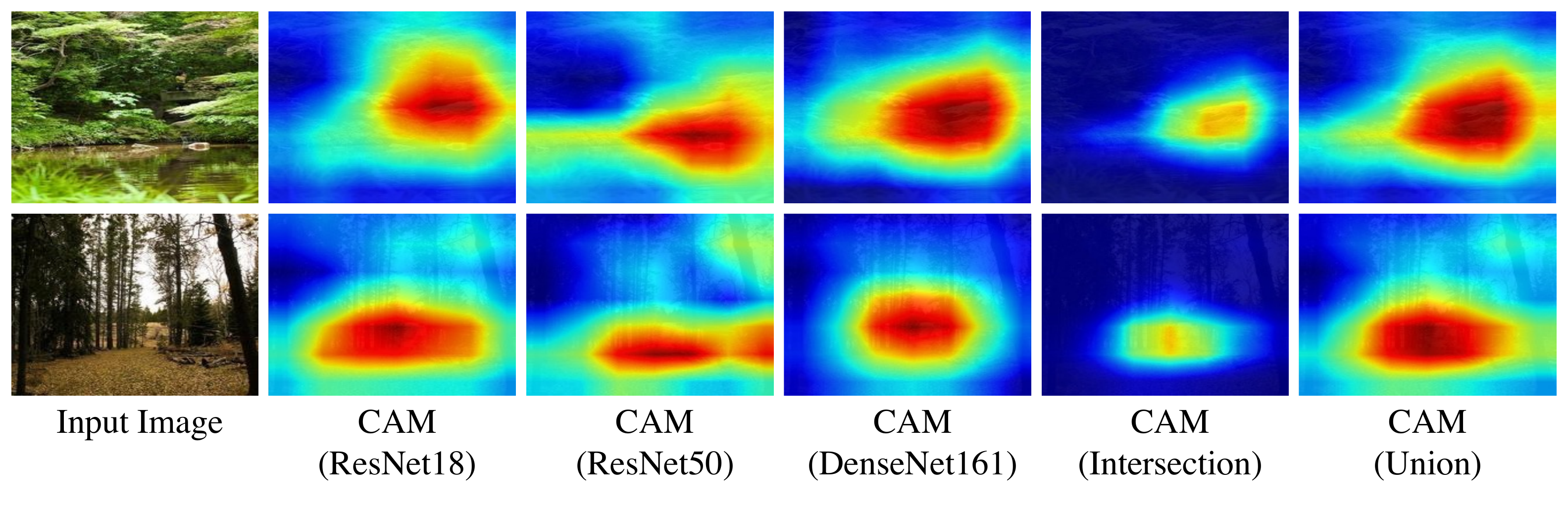}
    \caption{Examples of different protected region identification strategies using class activation map.}
    \label{fig:cam_examples}
\end{figure}

\subsection{Gradient-Based Adversarial Examples}

% This work considers only white-box attacks, which means the attacker has access to the model gradients \ie, the attacker has a copy of your model’s weights. This threat model gives the attacker much more power than black-box attacks. They can specifically craft their attack to fool your model without relying on transfer attacks that often result in human-visible perturbations.

Given an input image $x$ and a deep model $C_{\theta}$ with its parameters $\theta$, the \textbf{fast gradient sign method (FGSM)}~\cite{Goodfellow-ICLR2014}, an attack for an $\ell_\infty$-bounded adversary, computes an adversarial example:
\begin{equation}
    \widehat{x} = x + \epsilon \ast \text{sign} (\nabla_{x} \mathcal{L} ( x, y ; \theta )),
\end{equation}
where $\widehat{x}$ represents the adversarial result, $\epsilon$ denotes the magnitude of the perturbation, $\mathcal{L}$ represents the loss function, and $y$ is the correct label. $\nabla_{x}\mathcal{L} ( x, y ; \theta )$ is the gradient used to compute small adjustments of $x$ using loss function $\mathcal{L}$. 

FGSM performs only a one-step gradient update along the direction of the sign of the gradient at each pixel. In contrast, \textbf{projected gradient descent (PGD)}~\cite{Bubeck-arxiv2014}, which is also called the “iterative fast gradient sign method,” iterates by updating the gradient:
\begin{equation}
    \widehat{x}^{t+1} = \Pi( \widehat{x}^t + \epsilon \ast \text{sign} (\nabla_{x} \mathcal{L} ( \widehat{x}^t, y ; \theta ))),
\end{equation}
where $\widehat{x}^t$ is the adversarial result at the $t$-th step with $\widehat{x}^0 = x$; $\Pi$ represents the projection operator, which clips the input at positions in the predefined perturbation range. In this paper, we focus on non-Euclidean PGD, in which the $\ell_\infty$-norm is used as a distance function for deep classifiers. PGD initializes the example to a random point in the $\ell_\infty$-ball of interest, which is determined by the $\ell_\infty$-norm, and performs random restarts.

\subsection{Mask-Guided Multimodal Projected Gradient Descent}

\subsubsection{Multimodal Adversarial Example}

To improve the transferbility of adversarial examples, we extend PGD to multimodal PGD (M-PGD). In this algorithm, adversarial examples are trained on different deep models:
\begin{equation}
    \widehat{x}^{t+1} = \Pi ( \widehat{x}^t + \epsilon \ast \text{sign} (\nabla_{x} \Sigma_{k  \in K} \mathcal{L} ( \widehat{x}^t, y ; \theta_k ))),
\end{equation}
where $K$ denotes the number of deep models used. We trained M-PGD on three deep models: ResNet18~\cite{He-CVPR2016}, ResNet50~\cite{He-CVPR2016}, and DenseNet161~\cite{Huang-CVPR2017}. We also used models pre-trained on the Places365 dataset~\cite{Zhou-TPAMI2017}, which has been publicly released by the authors\footnote{\url{https://github.com/CSAILVision/places365}}. We empirically set $\epsilon = 0.03$ and $t \in \{ 1,..,20\}$ in all experiments, meaning that the M-PGD algorithm was iterated 20 times.

\begin{figure}[t!]
    \centering
    \includegraphics[width=\linewidth]{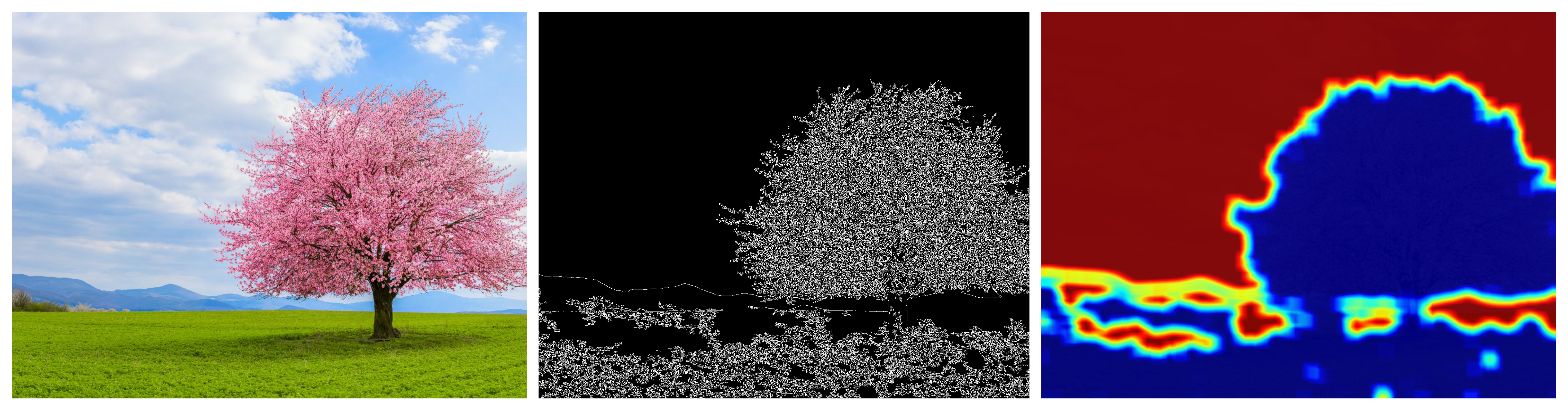}
    \caption{Left: original image. Middle: image texture, or image gradient, obtained using Canny edge detection algorithm. Right: protected regions, obtained by blending adversarial examples, which are less likely to attract to attention.}
    \label{fig:canny}
\end{figure}

\begin{figure*}[t!]
    \centering
    \includegraphics[width=\linewidth]{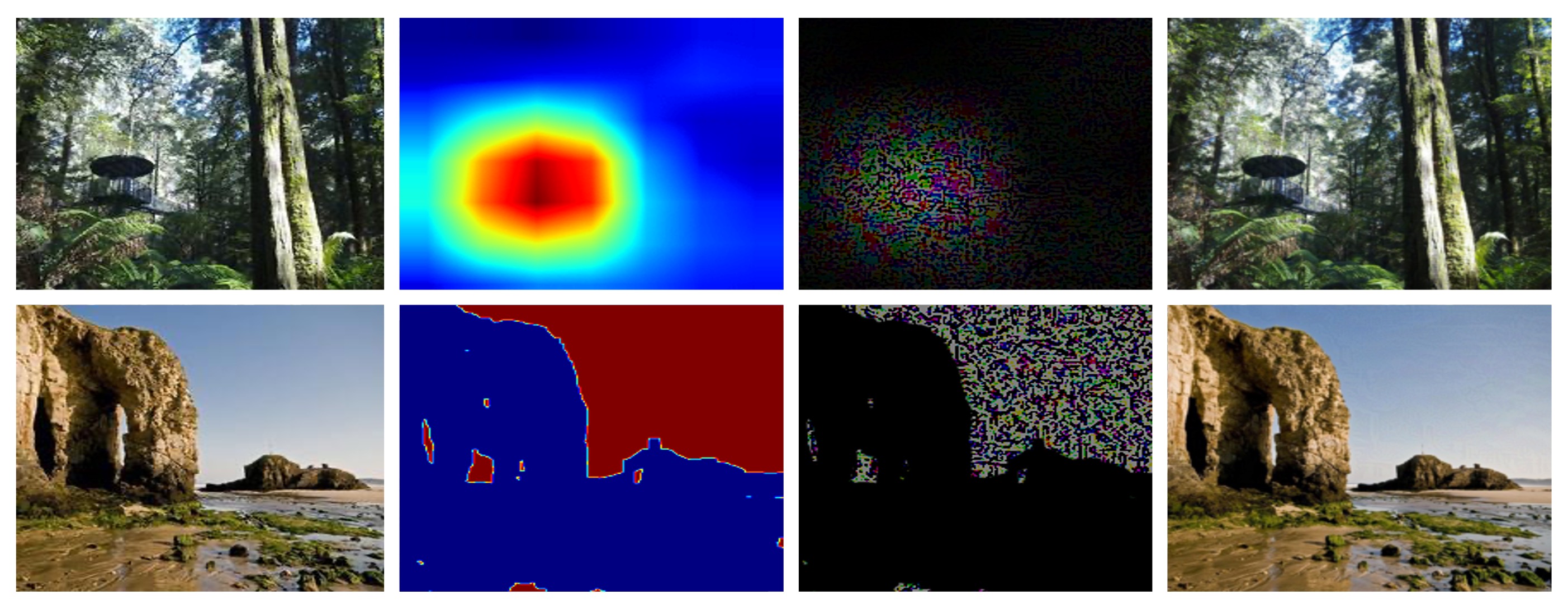}
    \caption{Examples of adversarial examples generated by CAM-based MM-PGD (top row) and HV-based MM-PGD (bottom row). From left to right, original images followed by masked protection regions, adversarial perturbations against ResNet50~\cite{He-CVPR2016}, and adversarial example results. Perturbation size is magnified 100 times for visualization purposes.}
    \label{fig:adversarial_examples}
\end{figure*}

\subsubsection{Mask-Guided Adversarial Example}

To investigate the properties of regions in images that need to be protected, we introduce mask-guided M-PGD (MM-PGD):
\begin{equation}
    \widehat{x}^{t+1} = \Pi ( \widehat{x}^t + \gamma_{att} \odot \epsilon \ast \text{sign} (\nabla_{x} \Sigma_{k  \in K} \mathcal{L} ( \widehat{x}^t, y ; \theta_k ))),
\end{equation}
where $\gamma_{att}$ denotes an attention map, which highlights image regions needing to be protected, and $\odot$ is the element-wise multiplication operator. We investigated attention maps in two directions: intrinsic and extrinsic essences. 

The intrinsic essence aims to exploit the image regions that reflect the internal behavior of CNNs. By protecting these regions by covering them with adversarial examples, we should be able to fool deep classifiers and thereby prevent them from exploiting private locations in the image (\cf~Fig. \ref{fig:adversarial_examples}). To achieve this goal, we use the CAM technique~\cite{Zhou-CVPR2016} to identify distinct image regions with particular properties (\cf~Fig.~\ref{fig:cam}). Indeed, a CAM lets us see which regions in the image are relevant to a specific class. Note that the final CAM is the union of the CAMs of all trained models (see Fig.~\ref{fig:cam_examples}).

The extrinsic essence aims to exploit image regions that are less likely to attract attention than other regions (\cf~Fig.~\ref{fig:adversarial_examples}). These regions are independent of trained models. To exploit these regions, we first detect image texture using a Canny edge detector~\cite{Canny-TPAMI1986}, followed by image processing operators. These regions tend to attract attention. We then take the inverse regions with the expectation that these regions are less likely to attract attention (\cf~Fig. \ref{fig:canny}).

\section{Experiments}
\label{sec:experiments}

\subsection{Experimental Settings}
\label{sec:settings}

% \subsubsection{Implementation} 

% \highlight{*** IMPORTANT: With current results, which are missed a lot, the possible way is using adv train on 3 models, and test on 1 model $=>$ ignore all other results in the slide because of conflicts.}

\subsubsection{Dataset} 

We used the Places365 dataset~\cite{Zhou-TPAMI2017}, which consists of 2.2 million images. From the test set, we randomly sampled 5000 images that could be recognized with $100\%$ accuracy by all deep classifiers used. 

\subsubsection{Evaluation Criteria} 

Given adversarial example method A that generates result $\widehat{x} = A(x)$ for an input $x$ with the corresponding label $y$ and a classifier $C$, the top-$k$ accuracy of the classifier against method A is defined as
\begin{equation}
    ACC_{k}(C,A)=\frac{1}{N}\sum_{i=1}^{N}{1(y_i \in C_{k}(A(x_i)))},
\end{equation}
where $\{x_i, y_i\}_{i=1}^{N}$ is the test set, with $N$ indicating the number of images in the test set; $1(\cdot)$ is the indicator function. $C_{k}(x)=\{\iota_j\}_{j=1}^{k}$ defines the top-$k$ output categories of the classifier with the conditions $\forall j \leq k$ and $k \leq K: p_C(\iota_j) \geq p(\iota_{j+1})$, where $K$ is the number of categories and $p_C(\iota)$ is the output probability of class $\iota$.

We define the protection rate (PR) to evaluate the top-$k$ robustness levels of untargeted adversarial example method A against classifier C as follows:
\begin{equation}
\resizebox{\linewidth}{!}{
    $PR_k(A,C)=\frac{1}{\sum_{i=1}^{N}1(y_i \in C(x_i))}\sum_{i=1}^{N}1(y_i \in C(x_i) \wedge y_i \notin C(A(x_i))).$
}
\end{equation}

We remark that the top-1 PR corresponds to the attack success rate (ASR), which is normally used to evaluate adversarial example methods designed for attack purposes. However, we used the top-$k$ PR with $k>1$ for protection evaluation. It is more suitable because the top-1 PR measures the number of the incorrect classifications obtained over the dataset, where the output label with the highest classification probability from the target model is incorrect. However, the output label with the second or third highest classification probability may be correct. Using the top-$k$ PR guarantees that all output labels with the $k$ highest classification probabilities are incorrect, which improves protection. We used the top-5 PR to evaluate the robustness of the adversarial example methods.

Since we sampled the data to ensure that all pre-trained deep classifiers used in the experiments achieved $100\%$ accuracy, as mentioned above, $y_i \in C_k(x_i)$ for all $\{x_i, y_i\}$ in the test set because $y_i = C_{k=1}(x_i)$. Hence, the PR we used can be simply reformulated as
\begin{equation}
    PR_k(A,C)=\frac{1}{N}\sum_{i=1}^{N}1(y_i \notin C(A(x_i)))=1-ACC_k(C,A).
\end{equation}

We used the structural similarity index measure (SSIM)~\cite{Wang-TIP2004} to quantify the adversarial examples.

% In addition, we used different metrics for quantifying adversarial examples, including Structural Similarity Index Measure (SSIM)~\cite{Wang-TIP2004} and Fréchet Inception Distance (FID)~\cite{Heusel-NeurIPS2017}. 

% https://openaccess.thecvf.com/content_CVPR_2020/papers/Dong_Benchmarking_Adversarial_Robustness_on_Image_Classification_CVPR_2020_paper.pdf

% \subsection{Ablation Studies}

\subsection{CAM-Based MM-PGD Evaluation}
\label{sec:experiments_1}

We demonstrated the usefulness of identifying the regions needing protection by using the CAM technique~\cite{Zhou-CVPR2016} with two different CAM combination strategies. The first strategy was to first generate CAMs using every trained deep model. We then took the intersection of the CAMs, denoted as ``Intersection-CAM." The second strategy was to take the union of CAMs, denoted as ``Union-CAM." Visualization examples are shown in Fig.~\ref{fig:cam_examples}. We also compared the performance of the methods with that of blending adversarial substitutions generated from M-PGD on entire images, denoted by ``Entire Image." The images without any protection from adversarial examples, defined as ``No Attack," had a PR of $0\%$, corresponding to a classification accuracy of $100\%$ and a SSIM of $100\%$?.

\begin{figure}[t!]
    \centering
    \includegraphics[width=\linewidth]{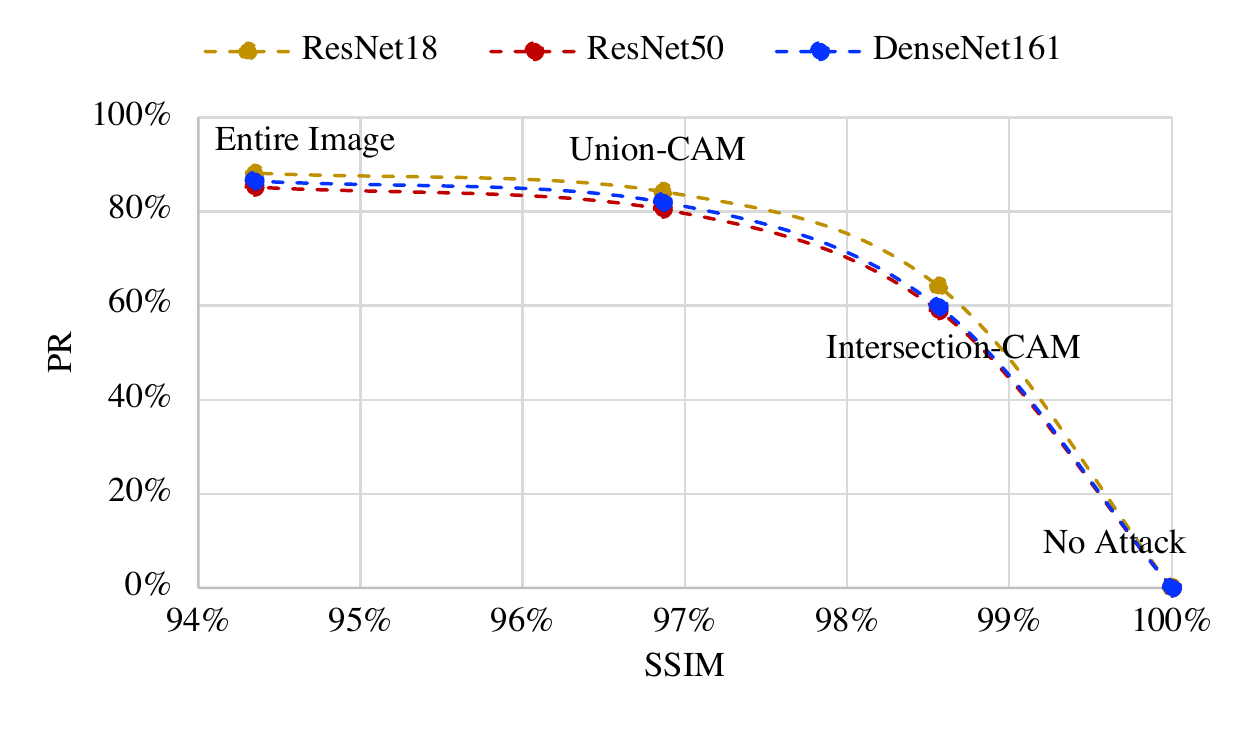}
    \caption{
    % Trade off between protection rate (PR) and image quality (SSIM). 
    Effectiveness of region identification using CAM. Utilizing union of CAMs achieved most balanced results: high PR against all three deep classifiers plus good image quality.}
    \label{fig:Exp_CAM}
\end{figure}

As shown in Fig.~\ref{fig:Exp_CAM}, the regions identified by the intersection of CAMs were too small, covering only about $25\%$ of the image pixels, and thus could not fully cover the regions of interest (ROIs). As a result, intersection-CAM had a PR of only about $60\%$, and its highest protection rate was $64.0\%$ against ResNet18. In contrast, the regions identified by the union of CAMs covered around $50\%$ of the image pixels. These regions fully covered the ROIs, resulting in a high PR (higher than $80\%$). Using blending adversarial examples on entire images improved the PR only slightly ($4\%$ higher than that of Union-CAM).

Hence, utilizing the union of CAMs can result in a high PR for images and good image quality. This strategy should be helpful in developing effective protection solutions against landmark recognition without requiring much image manipulation.

\subsection{HV-Based MM-PGD Evaluation}
\label{sec:experiments_2}

\begin{figure}[t!]
    \centering
    \includegraphics[width=\linewidth]{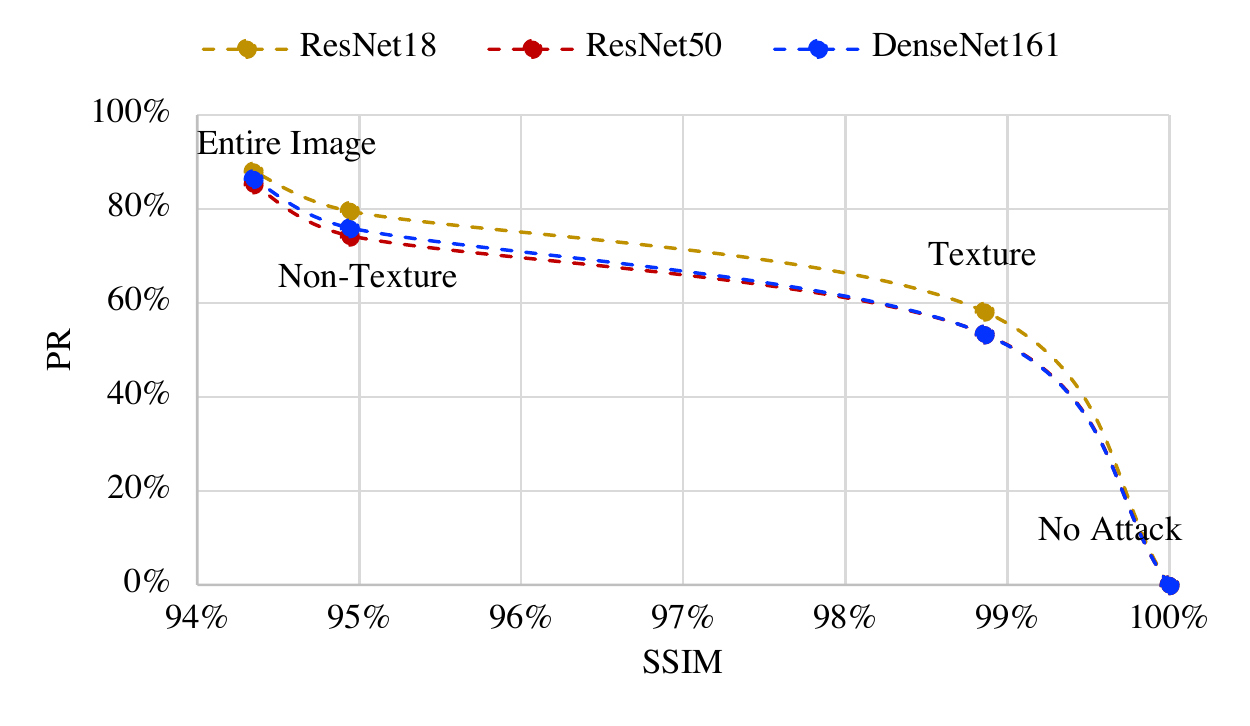}
    \caption{Effectiveness of region identification based on human vision. Targeting non-texture regions, which attract less attention, achieved high PR against all three deep classifiers while maintaining good image quality.}
    \label{fig:Exp_texture}
\end{figure}

In this section, we analyzed human-vision-based protected region identification. In particular, we compared the effectiveness of using texture and non-texture regions. 

Figure \ref{fig:Exp_texture} shows that applying adversarial examples in texture regions resulted in high transparency (SSIM of $99\%$). This is because these areas, which contain ROIs, account for a small percentage of the image area, leading to small protected areas. However, viewers usually focus on the ROIs in images and can recognize camouflaged artifacts (\ie, adversarial perturbations). Furthermore, small protected areas resulted in an unreasonable PR (less than $60\%$).

In contrast, applying blending adversarial examples in non-textured regions resulted in a high PR (about $80\%$). In terms of naturalness, these regions are usually background (\ie, sky, sea, ground, etc.) and thus do not contain ROIs. Therefore, viewers usually do not pay attention to these regions. Hence, we can blend adversarial examples into these regions and improve the strength of adversarial perturbations without affecting the image aesthetically.

\begin{figure}[t!]
    \centering
    \includegraphics[width=\linewidth]{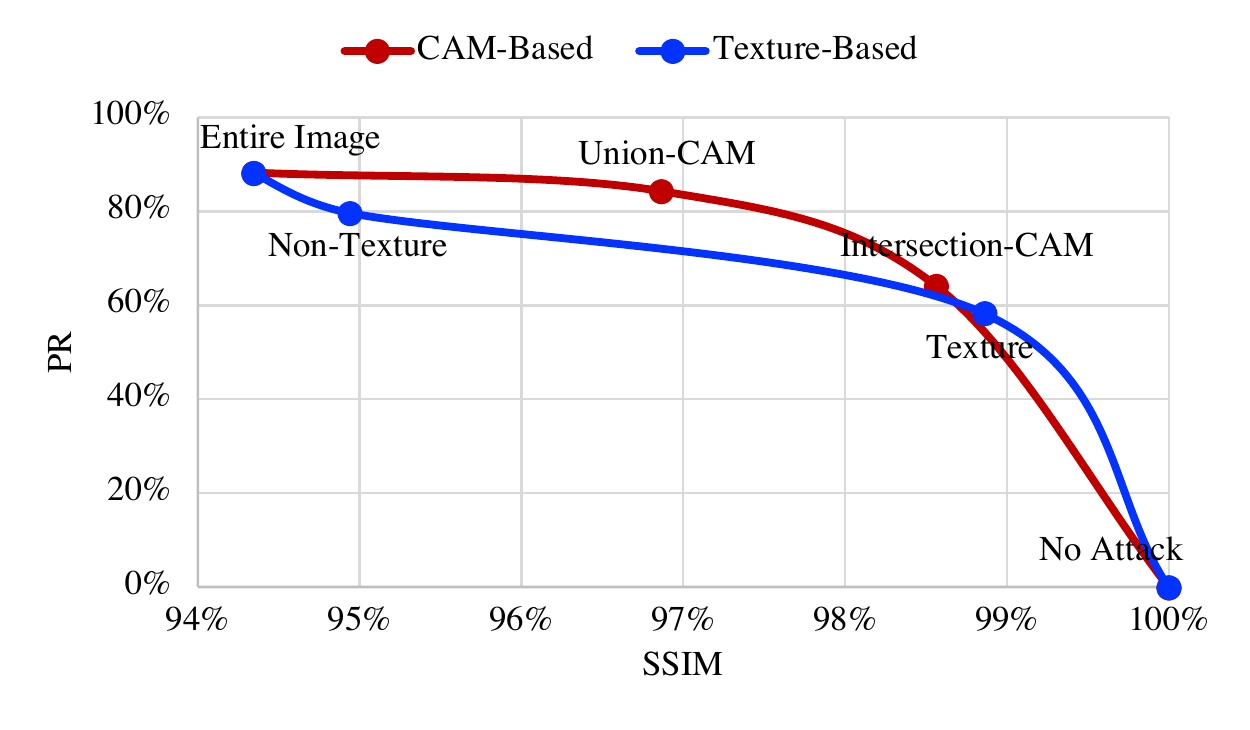}
    \caption{Trade-off between image quality and protection rate against ResNet18.}
    \label{fig:Exp_texture_cam}
\end{figure}

\begin{table}[t!]
\centering
\caption{Transferability of adversarial example methods against unknown WideResNet18 classifier.}
\label{table:transferability}
\begin{tabular}{|l|c|c|c|}
\hline
\textbf{Method} & \textbf{PR} $\Uparrow$ & \textbf{SSIM} $\Uparrow$  \\
\hline
Random noise & $2.6\%$ & $80.5\%$  \\
PGD-ResNet18 & $11.7\%$ & $95.0\%$  \\
PGD-ResNet50 & $10.7\%$ & $95.0\%$  \\
PGD-DenseNet161 & $10.9\%$ & $95.1\%$  \\
\rowcolor{lightgray} CAM-based MM-PGD & $20.1\%$ & $96.9\%$  \\
\rowcolor{lightgray} HV-based MM-PGD & $18.1\%$ & $94.9\%$  \\
\hline
\end{tabular}
\end{table}

We also investigated the advantages of our MM-PGD method by comparing CAM-based and HV-based region identification. It can be seen in Fig.~\ref{fig:Exp_texture_cam} that \textit{identifying regions on the basis of the union of CAMs has more advantages than applying adversarial examples in non-textured regions} in terms of image quality (SSIM) and PR. However, \textit{from a human vision viewpoint, blending adversarial examples into non-textured regions is more natural than blending them into CAM areas}. Indeed, as we stated above, MM-PGD, based on non-textured identification, targets background areas, which viewers tend to ignore. Thus, we can increase adversarial perturbation strength with a more negligible effect on the image aesthetic. In contrast, CAM-based MM-PGD focuses on regions that have meaning with deep models. These regions contain ROIs; thus, viewers usually pay attention to these areas and easily recognize artifacts if we blend in adversarial examples.

\subsection{Transferability Evaluation}

Our goal is to protect image contents from being exploited by landmark recognition systems, which are black-box APIs. Therefore, it is essential to investigate the performance of adversarial example methods against unknown deep classifiers that were not used in training the adversarial examples. To evaluate the transferability of our MM-PGD methods, we tested them on the WideResNet18~\cite{Zagoruyko-2016} model, which was not used in the training. We also used a model pre-trained on the Places365 dataset~\cite{Zhou-TPAMI2017} created by the authors\footnote{\url{https://github.com/CSAILVision/places365}}. 

We compared the performances of our MM-PGD methods trained on three deep models (\ie ResNet18, ResNet50, and DenseNet161) with those of ones trained on each model separately, denoted by “PGD-ResNet18,” “PGD-ResNet50,” and “PGD-DenseNet161.” Table \ref{table:transferability} shows that our multimodal training methods substantially outperformed those of ones with single-model training in terms of both PR and image quality. Indeed, the PR with single-model training was only around $11\%$, whereas the PRs of our MM-PGD methods were $20\%$ and $18\%$ for the CAM-based and HV-based methods, respectively. In addition, the image quality measured by SSIM was $95\%$, the same as with single-model training for the HV-based MM-PGD method and $97\%$ for the CAM-based MM-PGD method. Furthermore, as shown in the table, our methods had better performance than random noise protection, which was only $2.6\%$. These experimental results demonstrate the effectiveness of our proposed methods against black-box APIs.

\section{Conclusion}
\label{sec:conclusion}

We investigated a new application of adversarial examples, namely location privacy protection against deep landmark recognition systems. We extended the projected gradient descent (PGD) method to mask-guided multimodal PGD (MM-PGD) to investigate the transferability of adversarial examples. The adversarial examples were trained on three deep classifiers. We also investigated different approaches to protect regions: targeting deep model behaviors through class activation map (CAM)-based MM-PGD and targeting human vision via human vision (HV)-based MM-PGD. The experimental results suggest that these methods are suitable for defending against black-box landmark recognition systems without much image manipulation. We plan to investigate various factors of the given problem, for example, evaluating robustness against image compression and image processing and improving the generalization of our methods.

% https://www.kaggle.com/competitions/landmark-recognition-2021/overview
% https://www.kaggle.com/competitions/landmark-recognition-2019/code
% https://arxiv.org/pdf/2004.01804.pdf

% \pagebreak

\section*{Acknowledgements}
This work was partially supported by JSPS KAKENHI Grants JP16H06302, JP18H04120, JP21H04907, JP20K23355, and JP21K18023 and by JST CREST Grants JPMJCR18A6 and JPMJCR20D3, Japan.

% \pagebreak

% ---- Bibliography ----
%

% trigger a \newpage just before the given reference
% number - used to balance the columns on the last page
% adjust the value as needed - may need to be readjusted if
% the document is modified later
%\IEEEtriggeratref{8}
% The "triggered" command can be changed if desired:
%\IEEEtriggercmd{\enlargethispage{-5in}}

\bibliographystyle{IEEEtran}
\bibliography{short_bibtex}

\end{document}